\documentclass[conference]{IEEEtran}
\IEEEoverridecommandlockouts
\usepackage{cite}
\usepackage{amsmath,amssymb,amsfonts}
\usepackage{algorithmic}
\usepackage{graphicx}
\usepackage{textcomp}
\usepackage{xcolor}
\usepackage{multirow}
\usepackage{booktabs}
\usepackage{makecell}

\usepackage{amsmath,amsfonts,bm}

\def\eqref#1{equation~\ref{#1}}

\def\1{\bm{1}}

\def\ve{{\bm{e}}}

\def\vy{{\bm{y}}}

\def\mE{{\bm{E}}}

\DeclareMathAlphabet{\mathsfit}{\encodingdefault}{\sfdefault}{m}{sl}
\SetMathAlphabet{\mathsfit}{bold}{\encodingdefault}{\sfdefault}{bx}{n}

\usepackage{subfigure}

\makeatletter
\newcommand*{\centerfloat}{%
  \parindent \z@
  \leftskip \z@ \@plus 1fil \@minus \marginparwidth
  \rightskip \leftskip
  \parfillskip \z@skip}
\makeatother

\usepackage{tikz}
\usepackage[utf8]{inputenc}
\usepackage{pgfplots}
\usepackage{enumitem}
\DeclareUnicodeCharacter{2212}{−}
\usepgfplotslibrary{groupplots,dateplot}
\usetikzlibrary{patterns,shapes.arrows}
\pgfplotsset{compat=newest}

\newcommand{\quotes}[1]{``#1''}

\def \mainPlotHeight {6cm}
\def \plotHeightMiddle {6cm} 
\def \plotHeightSmall {6cm}  

\def\BibTeX{{\rm B\kern-.05em{\sc i\kern-.025em b}\kern-.08em
    T\kern-.1667em\lower.7ex\hbox{E}\kern-.125emX}}
\begin{document}

\title{Of Non-Linearity and Commutativity in BERT}


\author{\IEEEauthorblockN{Sumu Zhao}
\IEEEauthorblockA{
\textit{ETH Zurich, Switzerland }\\
Switzerland \\
zhaosumu@gmail.com}
\and
\IEEEauthorblockN{Dami\'an Pascual}
\IEEEauthorblockA{
\textit{ETH Zurich, Switzerland }\\
Switzerland \\
dpascual@ethz.ch}
\and
\IEEEauthorblockN{Gino Brunner}
\IEEEauthorblockA{
\textit{ETH Zurich, Switzerland }\\
Switzerland \\
brunnegi@ethz.ch}
\and
\IEEEauthorblockN{Roger Wattenhofer}
\IEEEauthorblockA{
\textit{ETH Zurich, Switzerland }\\
Switzerland \\
wattenhofer@ethz.ch}
}

\maketitle

\begin{abstract}
In this work we provide new insights into the transformer architecture, and in particular, its best-known variant, BERT. First, we propose a method to measure the degree of non-linearity of different elements of transformers. Next, we focus our investigation on the feed-forward networks (FFN) inside transformers, which contain two thirds of the model parameters and have so far not received much attention. We find that FFNs are an inefficient yet important architectural element and that they cannot simply be replaced by attention blocks without a degradation in performance. Moreover, we study the interactions between layers in BERT and show that, while the layers exhibit some hierarchical structure, they extract features in a fuzzy manner. Our results suggest that BERT has an inductive bias towards layer commutativity, which we find is mainly due to the skip connections. This provides a justification for the strong performance of recurrent and weight-shared transformer models.
\end{abstract}


\section{Introduction}
Since the introduction of the transformer architecture~\cite{vaswani2017attention}, and in particular of BERT~\cite{2019bert}, researchers have tried to gain a deeper understanding of the inner workings of these models \cite{bertology}. 
This research paves the way for better explanations of model decisions, thereby broadening the range of possible applications of transformers. 
Furthermore, based on works on understanding transformers, in \cite{long-short-attention-iclr2020} the authors introduce an improved self-attention mechanism, resulting in state of the art performance. This shows that deeper knowledge of transformers can not only help with interpretability, but also lead to new and better architectures. 
In this work, we aim to advance the understanding of the inner workings of transformers by providing novel insights from two complementary angles.

First, we assess the importance and representational power of different parts of BERT. We propose to use linear approximators to measure the degree of (non-)linearity of BERT's building blocks. Due to the large amounts of data used for pre-training, overfitting has so far not been an issue for transformers. This is exemplified by ever larger models achieving state of the art performance~\cite{t5, brown2020language}. It thus seems clear that models with higher expressive power are beneficial for natural language tasks. We therefore hypothesise that the degree of non-linearity of a transformer can give us valuable insight into its inner workings, and in particular, it can indicate which components are modeling more complex relationships.
Further, we run exhaustive experiments in which we remove or replace parts of the model and measure the impact on performance. We find that the feed-forward networks (FFN) in the transformer architecture are inefficient compared to the self-attention operation, especially since they make up two thirds of the model parameters. 

Second, we investigate why recent ``recurrent transformers'' such as ALBERT~\cite{Lan2020ALBERT}, which employs weight sharing across layers, performs on par with or even better than the original BERT. Here, we find that BERT has an inherent tendency towards weight-sharing, and that layers can be removed or swapped without causing much performance degradation. By comparing to simpler feed-forward architectures, we find that skip connections play an important role in inducing layer commutativity in neural networks.

Our main contributions are as follows:
\begin{itemize}
    \item A diagnosis tool to measure non-linearity in Transformers that takes into account the geometry of the embedding space.
    \item An extensive architectural study revealing the importance of the FFNs.
    \item An empirical justification for why BERT is amenable to weight-sharing across layers, indicating that skip connections are largely responsible for this.
\end{itemize}


\section{Related Work}

The field of Natural Language Processing (NLP) has undergone major developments since the introduction of the transformer architecture~\cite{vaswani2017attention}. Since then, many different variants of transformers have been proposed~\cite{GPT, GPT2, roberta, distilbert, t5, xlnet, reformer}. One of the fundamental building blocks of transformers is self-attention, which allows the model to learn alignments between words. The possibility of visualizing attention distributions and the quest for interpretability in deep learning have sparked a lot of research into understanding transformer models. 
Most of this research has focused on BERT~\cite{2019bert}, a well-known transformer model, initiating a line of work now known under the term BERTology~\cite{bertology}.

In this field, a number of studies look into attention distributions in order to understand transformers' decisions and their internal dynamics~\cite{clark2019does,htut2019attention,kovaleva2019revealing}. However, recent work raises concerns about the interpretability of attention distributions~\cite{OnIdentifiabilityInTransformers, pascual2020telling,pruthi2019learning,jain2019attention}. A complementary line of research uses probing classifiers to test for linguistic properties of different parts of the model~\cite{tenney2018what, lin2019open}. 

In this work, we take an architectural approach to the analysis of BERT and look into how different components of the model interact to extract features and process language. Similar to probing classifiers, we use a linear approximator to regress the output of different architectural components in order to study their non-linearity. While existing work has focused on analyzing the self-attention heads and the impact of model depth \cite{karthikeyan2020cross,voita2019analyzing,michel2019sixteen}, we direct our investigation towards the FFN blocks of BERT.  

The authors of \cite{tenney2019bert} and \cite{jawahar2019does} show that features are extracted in BERT following the same order as in the classical NLP pipeline, with syntactic features extracted in earlier layers and semantic features in later layers. These features are extracted in a fuzzy manner, in the sense that multiple layers are responsible for solving the same tasks (e.g., Part-Of-Speech (POS) tagging). We study this phenomenon from a complementary angle by investigating the similarity between layers. Remarkably, ALBERT~\cite{Lan2020ALBERT} directly shares the weights across the layers of BERT without loss of performance, and Universal Transformers~\cite{dehghani2018universal} use recurrence over one single transformer layer. Similarly, in \cite{press2020improving} the authors exploit layer similarity to re-order the layers and improve performance, while in~\cite{durrani2020analyzing} and~\cite{sajjad2020poor} they analyze layer redundancy to prune BERT with minimal performance degradation.
Our findings support the idea that BERT has an inductive bias towards weight sharing, which results in the layers having a strong degree of commutativity and the features being extracted in an incremental or fuzzy manner.

\section{Background}

\begin{figure}[t]
    \centering
    \subfigure[BERT]{
        \begin{minipage}{3.7cm}
            \centering
            \includegraphics[scale=0.9]{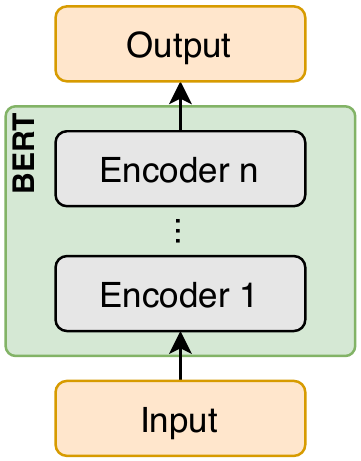}
        \end{minipage}}
    \subfigure[Encoder]{
        \begin{minipage}{3.7cm}
            \centering
            \includegraphics[scale=0.9]{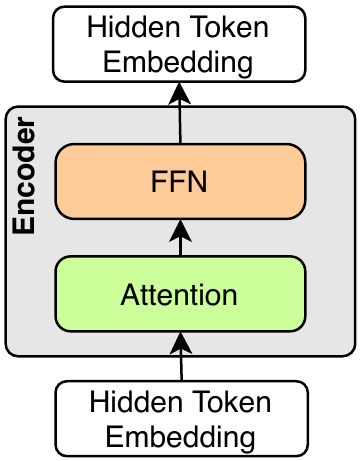}
        \end{minipage}}
    \caption{Architecture of BERT and the Encoder block. }
    \label{fig:bert-encoder}
\end{figure}

\begin{figure}[t]
    \centering
    \subfigure[Attention]{
        \begin{minipage}{3.7cm}
            \centering
            \includegraphics[scale=0.85]{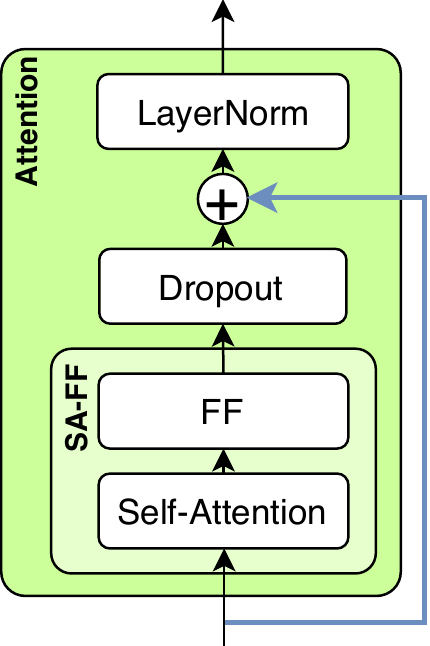}
        \end{minipage}}
    \subfigure[FFN]{
        \begin{minipage}{3.7cm}
            \centering
            \includegraphics[scale=0.85]{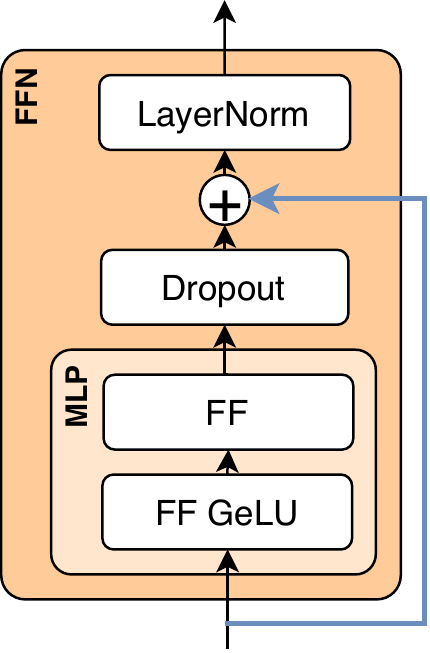}
        \end{minipage}}
    \caption{Architecture of Attention and FFN blocks.}
    \label{fig:attention-ffn}
\end{figure}

In this section, we give a brief description of BERT. For more details please refer to \cite{2019bert} and the official GitHub repository.\footnote{https://github.com/google-research/bert} 
As Figure~\ref{fig:bert-encoder} (a) shows, BERT is composed of $n$ encoder layers.
An encoder block follows the same structure as in the original transformer~\cite{vaswani2017attention}; it consists of an attention block and a feed-forward network (FFN), as shown in Figure~\ref{fig:bert-encoder} (b). Figure~\ref{fig:attention-ffn} shows the detailed structures of the attention and FFN blocks respectively. 
The input of BERT 
is a sequence of tokens, additionally containing special [CLS], [SEP] and [PAD] tokens. [CLS] can be used as output for classification tasks and [SEP] denotes sentence boundaries.
Since BERT is trained with a fixed (maximum) sequence length, [PAD] tokens have to be appended to the actual input sequences.

In our study, we use two BERT models of different sizes (base and small); architecture details are shown in Table~\ref{tbl:bert_model}. For BERT-base we use the pre-trained weights as released by the authors. All variants of BERT-small are pre-trained by us from scratch. Pre-training BERT-small takes 8 days on one Tesla V100 GPU card and 5 days on two cards. 
In \cite{2019bert} they pre-trained BERT-base on a 16GB text corpus which is a combination of E\footnotesize NGLISH\normalsize W\footnotesize IKIPEDIA \normalsize and B\footnotesize OOK\normalsize C\footnotesize ORPUS\normalsize~\cite{DBLP:conf/iccv/ZhuKZSUTF15}. Since this dataset is not readily available, we use O\footnotesize PEN\normalsize W\footnotesize EB\normalsize T\footnotesize EXT\normalsize~\cite{Gokaslan2019OpenWeb}, which is a 38GB text corpus created using the same method as for training GPT-2~\cite{GPT2}.

We evaluate our BERT-small models on all nine GLUE~\cite{Glue2019} tasks. For other experiments, we use the MNLI-matched (MNLIm) dataset from GLUE. To speed up the experiments, we select a random subset, MNLIm-sub, consisting of 39,271 examples for fine-tuning, and 983 for validation.

\begingroup
\setlength{\tabcolsep}{3pt} 
\begin{table}[t]
\small
\centering
\caption{Details of the BERT models we use.}
\begin{tabular}{llllll}
\toprule
\textbf{Model} & \textbf{Layers} & \textbf{Heads} & \makecell{ 
\textbf{Hidden}\\ \textbf{Size}}& \textbf{Params} & \makecell{ 
\textbf{Max. Seq}\\ \textbf{Length}} \\ \midrule
small & 6 & 8 & 512 & 35M & 128\\
base & 12 & 12 & 768 & 110M & 512\\
\bottomrule
\end{tabular}

\label{tbl:bert_model}
\end{table}
\endgroup

\section{Studying (non-)linearity}

A reliable measure of how non-linear different components of deep neural networks are, can help guide model analysis and design.
Here, we propose to use a linear approximator to quantify the non-linearity of different components of transformers. 

Given an architectural component $f$, we train a linear approximator $f^*$ that, given an input $\ve$, maximizes the cosine similarity between the actual output $f(\ve)=\vy$ and the approximated output $f^*(\ve) = \vy^*$. Cosine similarity is a common choice to measure similarity between words in an embedding space~\cite{pennington2014glove, cones_paper, OnIdentifiabilityInTransformers} . We define the linearity score $\gamma_f$ of a component $f$ as the average cosine similarity between actual and approximated outputs over a set of inputs: $\gamma_f = \mathbb{E}_{\ve \sim E}[\cos(\vy, \vy^*)]$. The larger $\gamma_f$, the more linear the measured component $f$ is. Intuitively, if $f$ is very non-linear, a linear approximator will not be able to approximate it accurately and the linearity score $\gamma_f$ will be close to $0$. If $f$ is very linear, $\gamma_f$ will be close to $1$.

\subsection{Method validation}

To validate our method, we first consider a simple toy task where we can manipulate the model's degree of non-linearity.
Specifically, we use a Multi-layer Perceptron (MLP) on the MNIST handwritten digit classification task, which consists of 60,000 examples. The MLP has one hidden layer of size 768 with ReLU activations, and one softmax layer with 10 output classes. 
We train this model on 50,000 examples for 30 epochs with batch size 32 and a learning rate of 0.001 using the Adam optimizer, and then evaluate it on 10,000 examples. 
Next, we use the remaining 10,000 training examples to train a linear model with 768 neurons for 10 epochs
to approximate the output activations of the MLP's hidden layer. 

\begin{figure}[t]
    \centerfloat
\begin{tikzpicture}

\begin{axis}[
legend cell align={left},
legend style={at={(1,0)}, anchor=south east, draw=white!80.0!black, font=\scriptsize},
tick align=outside,
tick pos=left,
height = \plotHeightSmall,
width = \linewidth,
x grid style={white!69.01960784313725!black},
xlabel={L2 Regularization Weight},
xmin=-0.45, xmax=9.45,
xtick style={color=black},
xtick={0,1,2,3,4,5,6,7,8,9},
xticklabels={0,1e-4,5e-4,1e-3,5e-3,1e-2,5e-2,1e-1,5e-1,1},
xticklabel style={font=\small, rotate=0},
y grid style={white!69.01960784313725!black},
ylabel={$\gamma$},
ymin=0.905, ymax=1.005,
ytick style={color=black}
]
\addplot [semithick, red, mark=square*, mark size=2, mark options={solid}]
table {%
0 0.999747
1 0.999729
2 0.999599
3 0.999521
4 0.999297
5 0.999071
6 0.998004
7 0.997122
8 0.998545
9 0.99948
};
\addlegendentry{Linear}
\addplot [semithick, black, mark=*, mark size=2, mark options={solid}]
table {%
0 0.911603
1 0.913682
2 0.925252
3 0.936349
4 0.963801
5 0.96808
6 0.984763
7 0.992184
8 0.99868
9 0.99949
};
\addlegendentry{Linear+ReLU}
\end{axis}

\end{tikzpicture}
    \caption{Linearity score of a hidden layer of a linear MLP (Linear) and a non-linear MLP (Linear+ReLU) for different regularization strengths.}
    \label{fig:cossim-mlp-l2reg-mnist}
\end{figure}
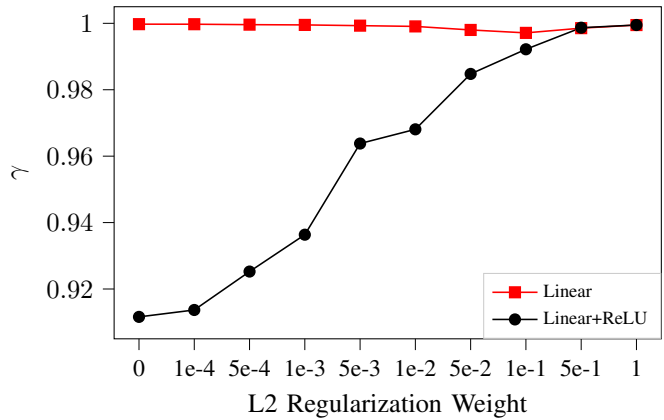

To verify the linearity score $\gamma$, we train the MLP with different L2 regularization strengths and calculate $\gamma$ for each model using the linear approximator.
We expect that the more non-linear a layer, the smaller $\gamma$, i.e., the larger the approximation error. Since increasing the regularization strength makes the weights smaller, which causes the hidden layer to behave more linearly, the linearity score $\gamma$ should increase with larger regularization weights. 
In Figure~\ref{fig:cossim-mlp-l2reg-mnist}, we observe this exact behavior, with a monotonic increase in linearity with stronger regularization. Next, we run the same experiment using a linear MLP (without ReLU activation). Since this model is linear, we expect a linearity score of $1$ regardless of the regularization strength; as shown in Figure~\ref{fig:cossim-mlp-l2reg-mnist}, we observe precisely this behavior.
These results show that the linear approximator can quantify the non-linearity of a simple MLP. 
Next, we apply this method to study the non-linearity of BERT. To do this, we first need to consider the geometry of the embedding space.

\begin{figure}[t]
\begin{tikzpicture}

\begin{axis}[
legend cell align={left},
legend columns = 3,
legend style={draw=white!80.0!black, at={(1, 1)},  font=\scriptsize},
tick align=outside,
tick pos=left,
height = \plotHeightSmall,
width = \linewidth,
x grid style={white!69.01960784313725!black},
xlabel={Layer},
xmin=-0.6, xmax=12.6,
xtick style={color=black},
xtick={0,1,2,3,4,5,6,7,8,9,10,11,12},
xticklabels={0,1,2,3,4,5,6,7,8,9,10,11,12}, 
y grid style={white!69.01960784313725!black},
ylabel={ConeSize},
ymin=0.04936, ymax=1,
ytick style={color=black}
]
\addplot [semithick, green!50.19607843137255!black, mark=*, mark size=2, mark options={solid}]
table {%
0 0.0884
1 0.1399
2 0.1993
3 0.1774
4 0.1975
5 0.2254
6 0.2181
7 0.2153
8 0.2125
9 0.1841
10 0.1843
11 0.2424
12 0.2281
};
\addlegendentry{Actual}
\addplot [semithick, blue, mark=square*, mark size=2, mark options={solid}]
table {%
0 0.7469
1 0.8692
2 0.8198
3 0.7888
4 0.7695
5 0.7114
6 0.6687
7 0.616
8 0.614
9 0.5967
10 0.6331
11 0.6875
12 0.5311
};
\addlegendentry{Pad}
\addplot [semithick, black, mark=triangle*, mark size=2, mark options={solid}]
table {%
0 0.3413
1 0.4861
2 0.5277
3 0.5173
4 0.5222
5 0.5137
6 0.4841
7 0.4493
8 0.4462
9 0.4217
10 0.4509
11 0.5216
12 0.3833
};
\addlegendentry{All}
\end{axis}

\end{tikzpicture}
    \caption{Average cosine similarity of 1000 random token pairs. The pairs are generated from the validation set of MNLIm-sub. Layer 0 denotes the input token embeddings, layers 1-12 denote the output token embeddings of the 12 layers of BERT. }
    \label{fig:baseline-cossim-mnlim}
\end{figure}
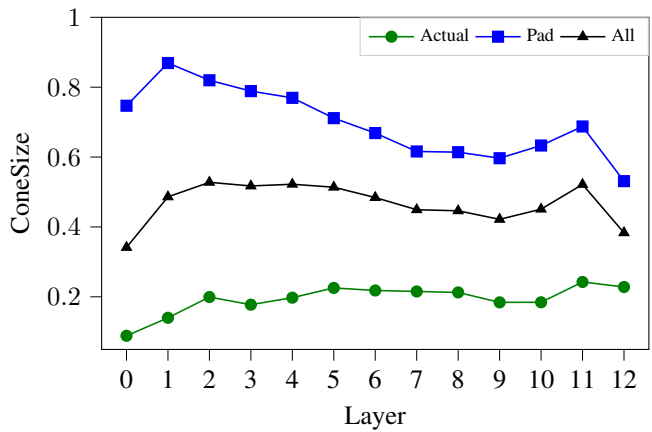

\subsection{Measuring the Non-Linearity of BERT}\label{subsec:measuring_nonlinearity}

In \cite{cones_paper} it is pointed out that the hidden embeddings at each transformer layer are directional and occupy cones of different sizes in their respective spaces, i.e., the embedding space is anisotropic, with most hidden embeddings located within a narrow region (cone) of the space. Since the size of these cones is different at each point of the model, we need to normalize the approximation errors by the respective cone sizes in order enable comparisons across layers. 
This is because, for architectural components that generate smaller cones at their output, the linear approximation is easier since all the output embeddings are already close to each other.

To calculate the cone size we follow \cite{cones_paper} and estimate the expected pair-wise cosine similarity from 1000 randomly sampled word representations. The intuition behind this is that the average similarity between pairs of randomly sampled vectors describes how narrow the cone is. This way, larger values correspond to smaller cones.
At each layer $l$, the cone size is defined as:
\begin{equation*}
    \text{ConeSize}(l) = \mathbb{E}_{(\ve_i^l,\ve_j^l) \sim \mE^l}[\cos(\ve_i^l,\ve_j^l)]
\end{equation*}
Where $\mE^l$ are the word embeddings at layer $l$. A value of $0$ represents perfect isotropy and $1$ perfect anisotropy.
We then calculate the cone size at each layer of BERT for the MNLIm-sub dataset. In Figure~\ref{fig:baseline-cossim-mnlim} we report the cone size for padding, actual and all tokens. We see that there is a significant difference between actual and padding tokens, and that the cone size for all tokens is roughly the average between both (weighted by the ratio of actual vs padding tokens). Overall, we see that the cone size for actual tokens is relatively wide\footnote{Our results differ from \cite{cones_paper}; we tried different settings without being able to reproduce the exact results. Possible explanations are that we separate padding and actual tokens, and that we use the original pre-trained weights of BERT while~\cite{cones_paper} uses the model from the HuggingFace~\cite{wolf2019transformers}.}.

This way, we calculate the cone size of the embeddings after different components $f$ in BERT. 
These values are then used to obtain the \emph{normalized} linearity score $\Tilde{\gamma}_f^l$ of component $f$ at layer $l$. 
To do this, we first subtract the cone size from $\gamma_f$ to correct for anisotropy, following the same procedure as in \cite{cones_paper}. Then, to allow comparison across layers, i.e., to have the values of $\Tilde{\gamma}_f^l$ contained in a fixed range, we normalize the result by dividing by the maximum possible value of the anisotropy-corrected linearity score $1-\text{ConeSize}(f_l)$: 
\begin{equation*}
    \Tilde{\gamma}_f^l = \frac{\gamma_f^l - \text{ConeSize}(f_l)}{1- \text{ConeSize}(f_l)}
\end{equation*}

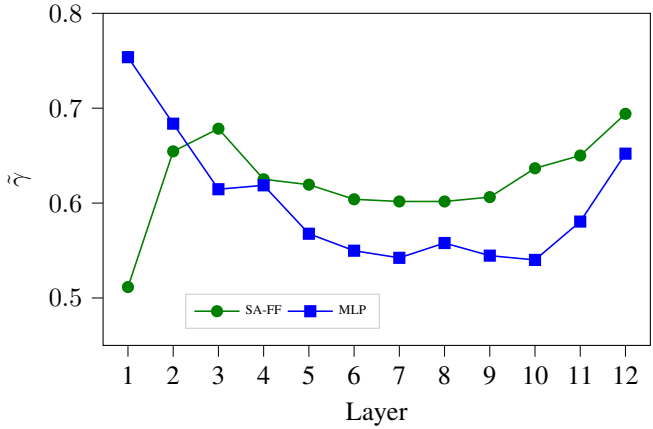
\begin{figure}[t]
\begin{tikzpicture}

\begin{axis}[
legend cell align={left},
legend style={at={(0.15,0.05)}, anchor=south west,
legend columns = 2,
draw=white!80.0!black, font=\tiny},
tick align=outside,
tick pos=left,
height = \plotHeightMiddle,
width = \linewidth,
x grid style={white!69.01960784313725!black},
xlabel={Layer},
xmin=-0.55, xmax=11.55,
xtick style={color=black},
xtick={0,1,2,3,4,5,6,7,8,9,10,11},
xticklabels={1,2,3,4,5,6,7,8,9,10,11,12},
y grid style={white!69.01960784313725!black},
ylabel={$\Tilde{\gamma}$},
ymin=0.45, ymax=0.8,
ytick style={color=black}
]


\addplot [semithick, green!50.19607843137255!black, mark=*, mark size=2, mark options={solid}]
table {%
0 0.5115
1 0.6545
2 0.6784
3 0.6251
4 0.6194
5 0.6040
6 0.6017
7 0.6017
8 0.6063
9 0.6367
10 0.6501
11 0.6941
};
\addlegendentry{SA-FF}


\addplot [semithick, blue, mark=square*, mark size=2, mark options={solid}]
table {%
0 0.7538
1 0.6837
2 0.6146
3 0.6187
4 0.5677
5 0.5498
6 0.5423
7 0.5579
8 0.5446
9 0.5401
10 0.5805
11 0.6521
};
\addlegendentry{MLP}


\end{axis}

\end{tikzpicture}
    \caption{Normalized linearity scores of SA-FF and MLP for each layer in BERT-base. The scores are obtained using MNLIm-sub.}
    \label{fig:cossim_base}
\end{figure}

Using this normalized linearity score, we measure at each layer of BERT the (non-)linearity of the SA-FF and the MLPs, which are the core of the Attention and FFN blocks respectively.
For both blocks, the linear approximator is a fully connected linear layer of the same size as the hidden dimension in BERT-base, i.e., 768. We train it for 3 epochs with batch size 64, using the Adam optimizer with a learning rate of 0.001 on MNLIm-sub. In Figure~\ref{fig:cossim_base} we show the normalized linearity score $\Tilde{\gamma}_f^l$ between the approximator output and the real tokens after each of the considered blocks for each of the layers of BERT-base.

Note that, while MLPs perform a non-linear function on one input token, the self-attention operation in the SA-FF block performs a weighted sum over all tokens in the input sequence. This way, the non-linearity introduced by SA-FF comes mainly from the aggregation of context information. Given the fundamentally different non-linear mechanisms used by each block, we refrain from directly comparing their linearity scores. Instead, we analyse how the non-linearity of each of them changes along the layers.

Both MLPs and SA-FF blocks follow opposing trends in the first three layers, with the MLPs becoming less linear, and SA-FF becoming more linear. After layer 3, the linearity scores of both blocks plateaus until approximately layer 10, when both become progressively more linear. 
Interestingly, the U-shaped pattern followed by both blocks (with the notable exception of the SA-FF block in the first layer), indicates that layers closer to the input/output space extract more linear features.

These results show that 1) MLPs introduce strong non-linearity, especially in the middle layers; and 2) the non-linearity introduced by SA-FF follows the same trend as that of MLPs even if the non-linear mechanism is fundamentally different. 
Next, we directly compare the impact of each block on model performance.

\subsection{MLP vs. Self-Attention}

Most research on understanding Transformer architectures has focused on the self-attention component. However, roughly two thirds of the total parameters of BERT reside within the FFN blocks, mainly in the MLPs. 
Here we compare the impact on performance of MLPs versus SA-FF blocks.

\begin{figure}[t]
    \centerfloat
\begin{tikzpicture}

\begin{axis}[
legend cell align={left},
legend style={at={(0.025,0.05)}, anchor=south west, draw=white!80.0!black, font=\scriptsize},
tick align=outside,
tick pos=left,
height = \plotHeightMiddle,
width = \linewidth,
x grid style={white!69.01960784313725!black},
xlabel={Layer},
xmin=-0.55, xmax=11.55,
xtick style={color=black},
xtick={0,1,2,3,4,5,6,7,8,9,10,11},
xticklabels={(12-12),(12-11),(12-10),(12-9),(12-8),(12-7),(12-6),(12-5),(12-4),(12-3),(12-2),(12-1)},
xticklabel style={font=\small, rotate=45},
y grid style={white!69.01960784313725!black},
ylabel={Accuracy},
ymin=0.36, ymax=0.82438,
ytick style={color=black}
]
\addplot [semithick, blue, dashed, mark=square*, mark size=2, mark options={solid}]
table {%
0 0.7831
1 0.7892
2 0.8004
3 0.776
4 0.7607
5 0.776
6 0.7627
7 0.72
8 0.7189
9 0.7149
10 0.6955
11 0.6701
};
\addlegendentry{Replace MLP}

\addplot [semithick, blue, mark=square*, mark size=2, mark options={solid}]
table {%
0 0.7994
1 0.7912
2 0.8035
3 0.7882
4 0.7627
5 0.7587
6 0.7342
7 0.7291
8 0.6965
9 0.6782
10 0.6884
11 0.6538
};
\addlegendentry{Remove MLP}


\addplot [semithick, green!50.19607843137255!black, mark=*, mark size=2, mark options={solid}]
table {%
0 0.7953
1 0.7912
2 0.779
3 0.7719
4 0.7515
5 0.7413
6 0.7016
7 0.6782
8 0.6721
9 0.6303
10 0.4878
11 0.4949
};
\addlegendentry{Remove SA-FF}
\addplot [semithick, black, mark=triangle*, mark size=2, mark options={solid}]
table {%
0 0.7882
1 0.7841
2 0.78
3 0.7627
4 0.7495
5 0.7301
6 0.7108
7 0.6701
8 0.6619
9 0.6008
10 0.5071
11 0.3859
};
\addlegendentry{Remove Encoder}
\end{axis}

\end{tikzpicture}
    \caption{Multi-layer backwards replacement/removal of different components in BERT-base, evaluated on MNLIm-sub.}
    \label{fig:backwards-replace-remove-mnlim}
\end{figure}
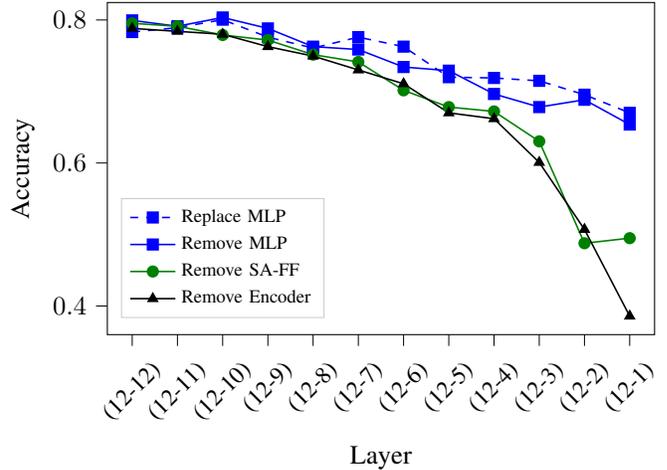

To this end, we first progressively replace the MLPs in all layers of BERT-base in the backwards direction, i.e., starting from the last layer, by their corresponding linear approximator. We replace backwards because this does not change the input to the earlier layers (which are not replaced). After replacement, we fine-tune and evaluate the model on MNLIm-sub. Then we repeat the process but this time we simply remove the MLPs, i.e., without replacing them by linear approximators. We compare the performance of these experiments with the performance of progressively removing the SA-FF blocks. The results are reported in Figure~\ref{fig:backwards-replace-remove-mnlim}; as reference, we also report the performance after removing the whole encoder layer.

Some surprising insights can be extracted from these experiments. First, replacing the MLPs by linear approximators yields approximately the same performance as completely removing the MLPs. This implies that it is the non-linear contribution of the MLP that is solely responsible for the performance difference. Second, the drop in performance after removing MLP blocks is smaller than when removing SA-FF blocks; especially after layer 6, removing the SA-FF blocks produces a much more severe degradation. Finally, removing SA-FF is as harmful as removing the whole encoder layer.

All of this confirms that while the majority of the parameters are in the MLPs, i.e., in the FFNs, the self-attention operation is more important for overall model performance. Furthermore, it raises the question of what happens if we remove all FFNs and instead add more attention layers. We explore this by training different variants of BERT-small and evaluating them on GLUE. In particular, we train: 

\begin{description}
    \item [no-FFN:] model without FFN blocks. This model has about one third of the original parameters. 
    \item [no-FFN-SA+]: model without FFN blocks, but with more layers so that the total number of parameters is roughly the same as BERT-small. This model has 18 layers.
    \item [no-FFN-GeLU-SA]: same as \emph{no-FFN} but with GeLU activation functions after the SA-FF block. 
    \item [no-FFN-GeLU-SA+]: same as \emph{no-FFN-SA+} but with GeLU activation functions after the SA-FF block. 
\end{description}

The results are shown in Table~\ref{tbl:glue_variants_small}. We find that, as expected, models with more parameters perform better; we see this from the performance of the \emph{SA+} models. We also see that adding a GeLU non-linearity~\cite{hendrycks2016gaussian} to the self-attention block further improves performance. However, adding more self-attention blocks in substitution of the FFN blocks degrades the performance of the model by more than 4 points, and by more than 2 points when using GeLU non-linearities. This strong degradation is surprising and indicates that separate FFN blocks with their own skip connection and layer normalization are more effective than moving the parameters and non-linearity to the attention blocks.

\begin{table}[t]
\small
\centering
\caption{GLUE scores of variants of BERT-small.}
\begin{tabular}{ll}
\toprule
\textbf{Model} & \textbf{GLUE score}  \\ \midrule
BERT-small & 74.5 \\
no-FFN & 69.6 \\ 
no-FFN-SA+ & 70.9 \\
no-FFN-GeLU-SA & 70.0 \\
no-FFN-GeLU-SA+ & 72.3 \\
\bottomrule
\end{tabular}
\label{tbl:glue_variants_small}
\end{table}

Overall, we find that the FFNs are an important, albeit somewhat inefficient, building block of BERT. We believe that a redesign of these structures could lead to improved performance, or equal performance with fewer parameters. However, this redesign is not straightforward and requires a more extensive investigation.

\section{BERT has an Inductive Bias Towards Weight Sharing}

Recently there has been a trend towards transformers that feature recurrence across the layer-dimension~\cite{Lan2020ALBERT,dehghani2018universal,DEQ}. 
For example, ALBERT achieves this by explicitly sharing the weights across all layers. 
Somewhat surprisingly, this does not negatively affect the performance, despite drastically reducing the number of parameters by a factor of $n_{layers}$.
In this section we show that BERT has an inductive bias that makes it amenable for explicit weight-sharing across layers. 
In particular, we show that the layers in BERT are highly commutative, that is, they can be swapped with surprisingly little impact on performance.

\subsection{Swapping Layers}

\begin{figure}[t]
    \centering
\begin{tikzpicture}

\begin{axis}[
legend cell align={left},
legend style={draw=white!80.0!black, anchor=south west, at={(0,0)}, font=\scriptsize},
tick align=outside,
tick pos=left,
height = \plotHeightSmall,
width = \linewidth,
x grid style={white!69.01960784313725!black},
xlabel={Swapped Layers},
xmin=-0.5, xmax=10.5,
xtick style={color=black},
xtick={0,1,2,3,4,5,6,7,8,9,10},
xticklabels={{(1,2)},{(2,3)},{(3,4)},{(4,5)},{(5,6)},{(6,7)},{(7,8)},{(8,9)},{(9,10)},{(10,11)},{(11,12)}},
xticklabel style={font=\small, rotate=45},
y grid style={white!69.01960784313725!black},
ylabel={Accuracy},
ymin=0.735835, ymax=0.812065,
ytick style={color=black}
]
\path [draw=black, semithick, dash pattern=on 5.550000000000001pt off 2.4000000000000004pt]
(axis cs:0,0.8014)
--(axis cs:10,0.8014);

\addplot [semithick, green!50.19607843137255!black, mark=*, mark size=2, mark options={solid}]
table {%
0 0.8035
1 0.7902
2 0.7851
3 0.7739
4 0.7943
5 0.8086
6 0.7719
7 0.7709
8 0.7393
9 0.7536
10 0.7912
};
\addlegendentry{Direct evaluation}
\addplot [semithick, blue, mark=square*, mark size=2, mark options={solid}]
table {%
0 0.7953
1 0.7882
2 0.7892
3 0.7862
4 0.7739
5 0.7923
6 0.7831
7 0.7851
8 0.7851
9 0.7862
10 0.7811
};
\addlegendentry{Train 3 more epochs}
\end{axis}

\end{tikzpicture}
    \caption{Swapping adjacient layers of BERT-base fine-tuned on MNLIm-sub. Baseline: 80.14\% (dashed).}
    \label{fig:swapping_two_bertbase}
\end{figure}
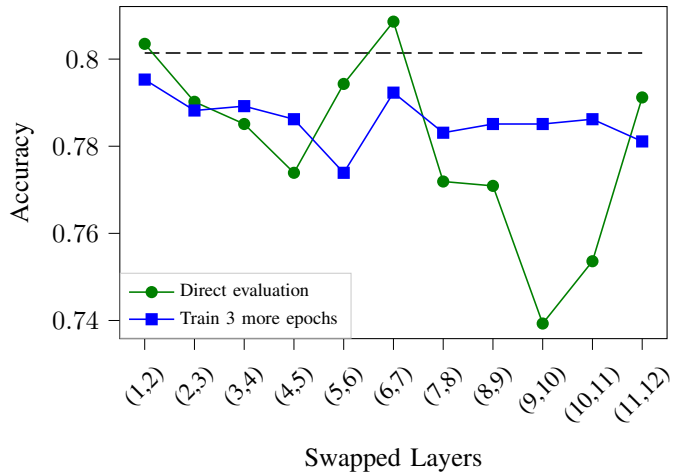

Figure~\ref{fig:swapping_two_bertbase} shows the effect of swapping \emph{adjacent} layers of BERT-base and evaluating the resulting models directly on the validation set of MNLIm-sub. Surprisingly, the average performance degradation is only 2\%. 

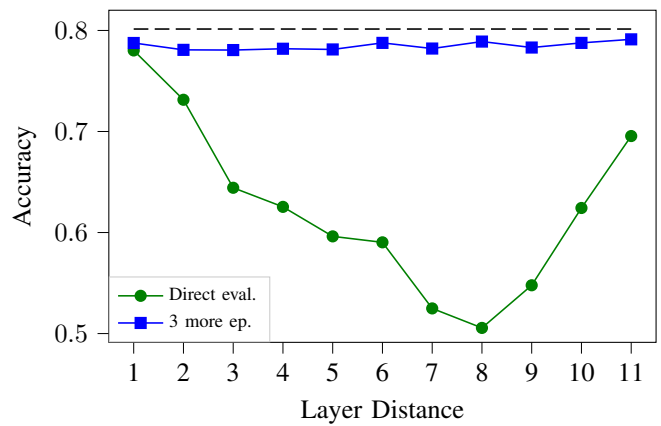
\begin{figure}[t]
    \centering
\begin{tikzpicture}

\begin{axis}[
legend cell align={left},
legend style={at={(0.,0.)}, anchor=south west, draw=white!80.0!black, font=\scriptsize},
tick align=outside,
tick pos=left,
height = \plotHeightSmall,
width = \linewidth,
x grid style={white!69.01960784313725!black},
xlabel={Layer Distance},
xmin=-0.5, xmax=10.5,
xtick style={color=black},
xtick={0,1,2,3,4,5,6,7,8,9,10},
xticklabels={1,2,3,4,5,6,7,8,9,10,11},
y grid style={white!69.01960784313725!black},
ylabel={Accuracy},
ymin=0.49132, ymax=0.82,
ytick style={color=black}
]
\path [draw=black, semithick, dash pattern=on 5.550000000000001pt off 2.4000000000000004pt]
(axis cs:0,0.8014)
--(axis cs:10,0.8014);

\addplot [semithick, green!50.19607843137255!black, mark=*, mark size=2, mark options={solid}]
table {%
0 0.7802
1 0.7314
2 0.6443
3 0.6254
4 0.5962
5 0.5903
6 0.5249
7 0.5056
8 0.5478
9 0.6243
10 0.6955
};
\addlegendentry{Direct eval.}
\addplot [semithick, blue, mark=square*, mark size=2, mark options={solid}]
table {%
0 0.7876
1 0.7808
2 0.7806
3 0.7819
4 0.7812
5 0.7877
6 0.7821
7 0.789
8 0.7831
9 0.7877
10 0.7912
};
\addlegendentry{ 3 more ep.}
\end{axis}

\end{tikzpicture}
    \caption{Swapping all possible combinations of two layers (mean accuracy). Baseline: 80.14\% (dashed).}
    \label{fig:swapping-bert-all-ditsances}
\end{figure}

Next, we investigate swapping \emph{any two} layers by evaluating all 66 models that can be created by swapping two layers. Figure~\ref{fig:swapping-bert-all-ditsances} shows the average accuracy for each swapping distance. Swapping distance refers to the absolute difference between the indices of the swapped layers. Note that there are fewer models for longer swapping distances and hence fewer data points. For example, swapping layers across a distance of 11 corresponds to swapping layer 1 and layer 12, and so there is only one possible model.
The figure shows that, overall, BERT is surprisingly robust to this modification, and only 3 epochs of further fine-tuning can recover most of the performance loss. However, swapping over longer distances generally results in worse performance, indicating that adjacent layers are more similar, as it was also observed in \cite{wu2020similarity}. Interestingly, for distances above 9 this trend reverses. When examining the results in more detail, we find that this is due to the fact that swapping layer 1 with either layer 11 or 12 results in accuracies of 62\% and 70\% respectively. This indicates that the very first and very last layers of BERT are remarkably similar, potentially due to their \quotes{closeness} to the input/output space. This is in line with the U-shape observed in our non-linearity analysis in Section~\ref{subsec:measuring_nonlinearity}.

\subsection{Shuffling Layers and ALBERT-like Configuration}

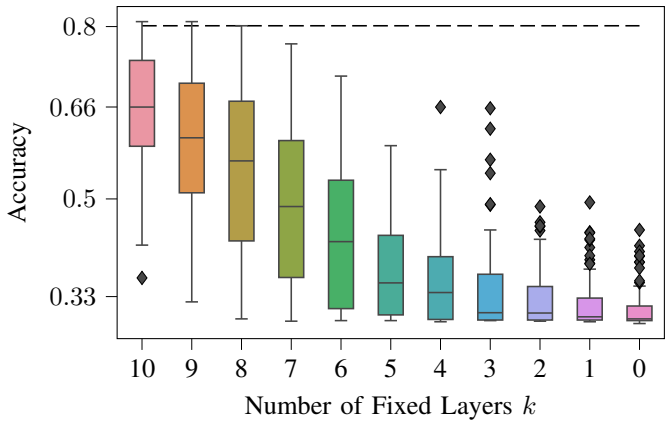
\begin{figure}[t]
    \centering
\begin{tikzpicture}

\definecolor{color0}{rgb}{0.916118471592631,0.587423760636183,0.640263770354658}
\definecolor{color1}{rgb}{0.863808356178602,0.573152210740463,0.307810812811634}
\definecolor{color2}{rgb}{0.700010791112321,0.614790479109793,0.283088087205159}
\definecolor{color3}{rgb}{0.557476983073822,0.647041308839097,0.274582905586777}
\definecolor{color4}{rgb}{0.283041109678183,0.688949210907788,0.403705480982841}
\definecolor{color5}{rgb}{0.291161607362895,0.676022094321277,0.595274339910988}
\definecolor{color6}{rgb}{0.300347004618729,0.669923443048919,0.691845126957855}
\definecolor{color7}{rgb}{0.330051495547042,0.675770945828907,0.827604419705628}
\definecolor{color8}{rgb}{0.664302243915371,0.674348614874934,0.921603226420326}
\definecolor{color9}{rgb}{0.842866822558884,0.584940072236322,0.909928522149295}
\definecolor{color10}{rgb}{0.909718848041829,0.562724016816337,0.786888997621333}

\begin{axis}[
tick align=outside,
tick pos=left,
height = \plotHeightMiddle,
width = \linewidth,
x grid style={white!69.0196078431373!black},
xlabel={Number of Fixed Layers $k$},
xmin=-0.5, xmax=10.5,
xtick style={color=black},
xtick={0,1,2,3,4,5,6,7,8,9,10},
xticklabels={10,9,8,7,6,5,4,3,2,1,0},
y grid style={white!69.0196078431373!black},
ylabel={Accuracy},
ymin=0.256825, ymax=0.834875,
ytick style={color=black},
ytick={0.33, 0.5, 0.66, 0.8}
]
\path [draw=black, semithick, dash pattern=on 5.550000000000001pt off 2.4000000000000004pt]
(axis cs:0,0.8014)
--(axis cs:10,0.8014);

\path [draw=white!27.843137254902!black, fill=color0, semithick]
(axis cs:-0.25,0.591675)
--(axis cs:0.25,0.591675)
--(axis cs:0.25,0.741075)
--(axis cs:-0.25,0.741075)
--(axis cs:-0.25,0.591675)
--cycle;
\path [draw=white!27.843137254902!black, fill=color1, semithick]
(axis cs:0.75,0.510675)
--(axis cs:1.25,0.510675)
--(axis cs:1.25,0.701375)
--(axis cs:0.75,0.701375)
--(axis cs:0.75,0.510675)
--cycle;
\path [draw=white!27.843137254902!black, fill=color2, semithick]
(axis cs:1.75,0.426925)
--(axis cs:2.25,0.426925)
--(axis cs:2.25,0.67005)
--(axis cs:1.75,0.67005)
--(axis cs:1.75,0.426925)
--cycle;
\path [draw=white!27.843137254902!black, fill=color3, semithick]
(axis cs:2.75,0.36325)
--(axis cs:3.25,0.36325)
--(axis cs:3.25,0.601575)
--(axis cs:2.75,0.601575)
--(axis cs:2.75,0.36325)
--cycle;
\path [draw=white!27.843137254902!black, fill=color4, semithick]
(axis cs:3.75,0.309075)
--(axis cs:4.25,0.309075)
--(axis cs:4.25,0.53255)
--(axis cs:3.75,0.53255)
--(axis cs:3.75,0.309075)
--cycle;
\path [draw=white!27.843137254902!black, fill=color5, semithick]
(axis cs:4.75,0.29815)
--(axis cs:5.25,0.29815)
--(axis cs:5.25,0.4366)
--(axis cs:4.75,0.4366)
--(axis cs:4.75,0.29815)
--cycle;
\path [draw=white!27.843137254902!black, fill=color6, semithick]
(axis cs:5.75,0.2902)
--(axis cs:6.25,0.2902)
--(axis cs:6.25,0.39945)
--(axis cs:5.75,0.39945)
--(axis cs:5.75,0.2902)
--cycle;
\path [draw=white!27.843137254902!black, fill=color7, semithick]
(axis cs:6.75,0.2892)
--(axis cs:7.25,0.2892)
--(axis cs:7.25,0.368875)
--(axis cs:6.75,0.368875)
--(axis cs:6.75,0.2892)
--cycle;
\path [draw=white!27.843137254902!black, fill=color8, semithick]
(axis cs:7.75,0.2892)
--(axis cs:8.25,0.2892)
--(axis cs:8.25,0.34755)
--(axis cs:7.75,0.34755)
--(axis cs:7.75,0.2892)
--cycle;
\path [draw=white!27.843137254902!black, fill=color9, semithick]
(axis cs:8.75,0.2892)
--(axis cs:9.25,0.2892)
--(axis cs:9.25,0.3274)
--(axis cs:8.75,0.3274)
--(axis cs:8.75,0.2892)
--cycle;
\path [draw=white!27.843137254902!black, fill=color10, semithick]
(axis cs:9.75,0.2882)
--(axis cs:10.25,0.2882)
--(axis cs:10.25,0.3136)
--(axis cs:9.75,0.3136)
--(axis cs:9.75,0.2882)
--cycle;
\addplot [semithick, white!27.843137254902!black]
table {%
0 0.591675
0 0.4196
};
\addplot [semithick, white!27.843137254902!black]
table {%
0 0.741075
0 0.8086
};
\addplot [semithick, white!27.843137254902!black]
table {%
-0.125 0.4196
0.125 0.4196
};
\addplot [semithick, white!27.843137254902!black]
table {%
-0.125 0.8086
0.125 0.8086
};
\addplot [black, mark=diamond*, mark size=2.5, mark options={solid,fill=white!27.843137254902!black}, only marks]
table {%
0 0.3625
};
\addplot [semithick, white!27.843137254902!black]
table {%
1 0.510675
1 0.3208
};
\addplot [semithick, white!27.843137254902!black]
table {%
1 0.701375
1 0.8086
};
\addplot [semithick, white!27.843137254902!black]
table {%
0.875 0.3208
1.125 0.3208
};
\addplot [semithick, white!27.843137254902!black]
table {%
0.875 0.8086
1.125 0.8086
};
\addplot [semithick, white!27.843137254902!black]
table {%
2 0.426925
2 0.2912
};
\addplot [semithick, white!27.843137254902!black]
table {%
2 0.67005
2 0.8014
};
\addplot [semithick, white!27.843137254902!black]
table {%
1.875 0.2912
2.125 0.2912
};
\addplot [semithick, white!27.843137254902!black]
table {%
1.875 0.8014
2.125 0.8014
};
\addplot [semithick, white!27.843137254902!black]
table {%
3 0.36325
3 0.2872
};
\addplot [semithick, white!27.843137254902!black]
table {%
3 0.601575
3 0.7699
};
\addplot [semithick, white!27.843137254902!black]
table {%
2.875 0.2872
3.125 0.2872
};
\addplot [semithick, white!27.843137254902!black]
table {%
2.875 0.7699
3.125 0.7699
};
\addplot [semithick, white!27.843137254902!black]
table {%
4 0.309075
4 0.2882
};
\addplot [semithick, white!27.843137254902!black]
table {%
4 0.53255
4 0.7138
};
\addplot [semithick, white!27.843137254902!black]
table {%
3.875 0.2882
4.125 0.2882
};
\addplot [semithick, white!27.843137254902!black]
table {%
3.875 0.7138
4.125 0.7138
};
\addplot [semithick, white!27.843137254902!black]
table {%
5 0.29815
5 0.2882
};
\addplot [semithick, white!27.843137254902!black]
table {%
5 0.4366
5 0.5927
};
\addplot [semithick, white!27.843137254902!black]
table {%
4.875 0.2882
5.125 0.2882
};
\addplot [semithick, white!27.843137254902!black]
table {%
4.875 0.5927
5.125 0.5927
};
\addplot [semithick, white!27.843137254902!black]
table {%
6 0.2902
6 0.2862
};
\addplot [semithick, white!27.843137254902!black]
table {%
6 0.39945
6 0.5509
};
\addplot [semithick, white!27.843137254902!black]
table {%
5.875 0.2862
6.125 0.2862
};
\addplot [semithick, white!27.843137254902!black]
table {%
5.875 0.5509
6.125 0.5509
};
\addplot [black, mark=diamond*, mark size=2.5, mark options={solid,fill=white!27.843137254902!black}, only marks]
table {%
6 0.6599
};
\addplot [semithick, white!27.843137254902!black]
table {%
7 0.2892
7 0.2882
};
\addplot [semithick, white!27.843137254902!black]
table {%
7 0.368875
7 0.446
};
\addplot [semithick, white!27.843137254902!black]
table {%
6.875 0.2882
7.125 0.2882
};
\addplot [semithick, white!27.843137254902!black]
table {%
6.875 0.446
7.125 0.446
};
\addplot [black, mark=diamond*, mark size=2.5, mark options={solid,fill=white!27.843137254902!black}, only marks]
table {%
7 0.5448
7 0.4898
7 0.6578
7 0.6222
7 0.5682
7 0.4908
};
\addplot [semithick, white!27.843137254902!black]
table {%
8 0.2892
8 0.2872
};
\addplot [semithick, white!27.843137254902!black]
table {%
8 0.34755
8 0.4297
};
\addplot [semithick, white!27.843137254902!black]
table {%
7.875 0.2872
8.125 0.2872
};
\addplot [semithick, white!27.843137254902!black]
table {%
7.875 0.4297
8.125 0.4297
};
\addplot [black, mark=diamond*, mark size=2.5, mark options={solid,fill=white!27.843137254902!black}, only marks]
table {%
8 0.4593
8 0.445
8 0.4521
8 0.4532
8 0.4868
};
\addplot [semithick, white!27.843137254902!black]
table {%
9 0.2892
9 0.2862
};
\addplot [semithick, white!27.843137254902!black]
table {%
9 0.3274
9 0.3778
};
\addplot [semithick, white!27.843137254902!black]
table {%
8.875 0.2862
9.125 0.2862
};
\addplot [semithick, white!27.843137254902!black]
table {%
8.875 0.3778
9.125 0.3778
};
\addplot [black, mark=diamond*, mark size=2.5, mark options={solid,fill=white!27.843137254902!black}, only marks]
table {%
9 0.4155
9 0.3941
9 0.4287
9 0.442
9 0.388
9 0.4939
9 0.4012
9 0.387
9 0.4409
9 0.4308
};
\addplot [semithick, white!27.843137254902!black]
table {%
10 0.2882
10 0.2831
};
\addplot [semithick, white!27.843137254902!black]
table {%
10 0.3136
10 0.3483
};
\addplot [semithick, white!27.843137254902!black]
table {%
9.875 0.2831
10.125 0.2831
};
\addplot [semithick, white!27.843137254902!black]
table {%
9.875 0.3483
10.125 0.3483
};
\addplot [black, mark=diamond*, mark size=2.5, mark options={solid,fill=white!27.843137254902!black}, only marks]
table {%
10 0.39
10 0.3544
10 0.4185
10 0.4002
10 0.39
10 0.4084
10 0.4012
10 0.3798
10 0.3574
10 0.446
};
\addplot [semithick, white!27.843137254902!black]
table {%
-0.25 0.6599
0.25 0.6599
};
\addplot [semithick, white!27.843137254902!black]
table {%
0.75 0.6064
1.25 0.6064
};
\addplot [semithick, white!27.843137254902!black]
table {%
1.75 0.5662
2.25 0.5662
};
\addplot [semithick, white!27.843137254902!black]
table {%
2.75 0.48675
3.25 0.48675
};
\addplot [semithick, white!27.843137254902!black]
table {%
3.75 0.42565
4.25 0.42565
};
\addplot [semithick, white!27.843137254902!black]
table {%
4.75 0.35385
5.25 0.35385
};
\addplot [semithick, white!27.843137254902!black]
table {%
5.75 0.33705
6.25 0.33705
};
\addplot [semithick, white!27.843137254902!black]
table {%
6.75 0.30195
7.25 0.30195
};
\addplot [semithick, white!27.843137254902!black]
table {%
7.75 0.3014
8.25 0.3014
};
\addplot [semithick, white!27.843137254902!black]
table {%
8.75 0.2948
9.25 0.2948
};
\addplot [semithick, white!27.843137254902!black]
table {%
9.75 0.2912
10.25 0.2912
};
\end{axis}

\end{tikzpicture}
    \caption{Create new BERT models by keeping $k$ layers fixed and shuffling the rest. The models are then directly evaluated on MNLIm-sub without further fine-tuning. Baseline: 80.14\% (dashed).}
    \label{fig:shuffle-direct-eval-fixklayer-100random}
\end{figure}

Next, we extend our analysis from swapping two layers at a time to shuffling different numbers of layers, i.e., changing their order.
Figure~\ref{fig:shuffle-direct-eval-fixklayer-100random} shows the results of a series of layer-shuffling experiments on BERT; we first randomly choose $k$ layers to be fixed, and then randomly permute all the remaining layers. For every value of $k$ we create 100 permuted versions of BERT. We then directly evaluate each of these models on the MNLIm-sub validation set without any additional fine-tuning. The results show that as we decrease the number of fixed layers, performance monotonically decreases. However, only for $k<=3$ does the median accuracy drops to random guessing (MNLI has three classes, thus random guessing corresponds to an accuracy of 33\%). Also, for every value of $k$, there are many models that still perform well above random. 

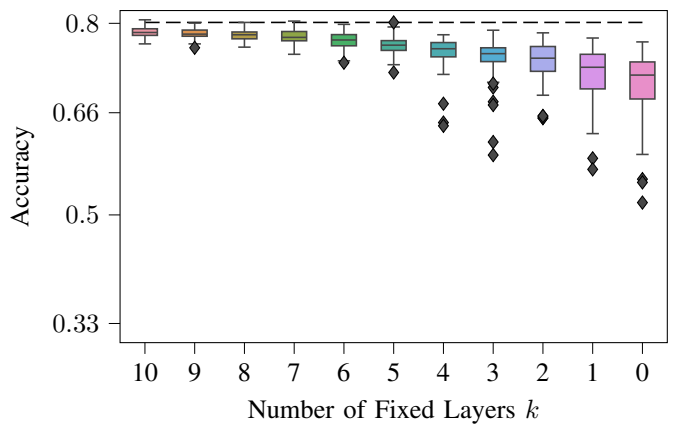
\begin{figure}[t]
    \centering
\begin{tikzpicture}

\definecolor{color0}{rgb}{0.916118471592631,0.587423760636183,0.640263770354658}
\definecolor{color1}{rgb}{0.863808356178602,0.573152210740463,0.307810812811634}
\definecolor{color2}{rgb}{0.700010791112321,0.614790479109793,0.283088087205159}
\definecolor{color3}{rgb}{0.557476983073822,0.647041308839097,0.274582905586777}
\definecolor{color4}{rgb}{0.283041109678183,0.688949210907788,0.403705480982841}
\definecolor{color5}{rgb}{0.291161607362895,0.676022094321277,0.595274339910988}
\definecolor{color6}{rgb}{0.300347004618729,0.669923443048919,0.691845126957855}
\definecolor{color7}{rgb}{0.330051495547042,0.675770945828907,0.827604419705628}
\definecolor{color8}{rgb}{0.664302243915371,0.674348614874934,0.921603226420326}
\definecolor{color9}{rgb}{0.842866822558884,0.584940072236322,0.909928522149295}
\definecolor{color10}{rgb}{0.909718848041829,0.562724016816337,0.786888997621333}

\begin{axis}[
tick align=outside,
tick pos=left,
height = \plotHeightMiddle,
width = \linewidth,
x grid style={white!69.0196078431373!black},
xlabel={Number of Fixed Layers $k$},
xmin=-0.5, xmax=10.5,
xtick style={color=black},
xtick={0,1,2,3,4,5,6,7,8,9,10},
xticklabels={10,9,8,7,6,5,4,3,2,1,0},
y grid style={white!69.0196078431373!black},
ylabel={Accuracy},
ymin=0.3, ymax=0.82,
ytick={0.33,0.5,0.66,0.8},
ytick style={color=black}
]
\path [draw=black, semithick, dash pattern=on 5.550000000000001pt off 2.4000000000000004pt]
(axis cs:0,0.8014)
--(axis cs:10,0.8014);

\path [draw=white!27.843137254902!black, fill=color0, semithick]
(axis cs:-0.25,0.7811)
--(axis cs:0.25,0.7811)
--(axis cs:0.25,0.791475)
--(axis cs:-0.25,0.791475)
--(axis cs:-0.25,0.7811)
--cycle;
\path [draw=white!27.843137254902!black, fill=color1, semithick]
(axis cs:0.75,0.77975)
--(axis cs:1.25,0.77975)
--(axis cs:1.25,0.78945)
--(axis cs:0.75,0.78945)
--(axis cs:0.75,0.77975)
--cycle;
\path [draw=white!27.843137254902!black, fill=color2, semithick]
(axis cs:1.75,0.775725)
--(axis cs:2.25,0.775725)
--(axis cs:2.25,0.78645)
--(axis cs:1.75,0.78645)
--(axis cs:1.75,0.775725)
--cycle;
\path [draw=white!27.843137254902!black, fill=color3, semithick]
(axis cs:2.75,0.77265)
--(axis cs:3.25,0.77265)
--(axis cs:3.25,0.7872)
--(axis cs:2.75,0.7872)
--(axis cs:2.75,0.77265)
--cycle;
\path [draw=white!27.843137254902!black, fill=color4, semithick]
(axis cs:3.75,0.7648)
--(axis cs:4.25,0.7648)
--(axis cs:4.25,0.78235)
--(axis cs:3.75,0.78235)
--(axis cs:3.75,0.7648)
--cycle;
\path [draw=white!27.843137254902!black, fill=color5, semithick]
(axis cs:4.75,0.7576)
--(axis cs:5.25,0.7576)
--(axis cs:5.25,0.7729)
--(axis cs:4.75,0.7729)
--(axis cs:4.75,0.7576)
--cycle;
\path [draw=white!27.843137254902!black, fill=color6, semithick]
(axis cs:5.75,0.7475)
--(axis cs:6.25,0.7475)
--(axis cs:6.25,0.7699)
--(axis cs:5.75,0.7699)
--(axis cs:5.75,0.7475)
--cycle;
\path [draw=white!27.843137254902!black, fill=color7, semithick]
(axis cs:6.75,0.74005)
--(axis cs:7.25,0.74005)
--(axis cs:7.25,0.7617)
--(axis cs:6.75,0.7617)
--(axis cs:6.75,0.74005)
--cycle;
\path [draw=white!27.843137254902!black, fill=color8, semithick]
(axis cs:7.75,0.724825)
--(axis cs:8.25,0.724825)
--(axis cs:8.25,0.7637)
--(axis cs:7.75,0.7637)
--(axis cs:7.75,0.724825)
--cycle;
\path [draw=white!27.843137254902!black, fill=color9, semithick]
(axis cs:8.75,0.697325)
--(axis cs:9.25,0.697325)
--(axis cs:9.25,0.7515)
--(axis cs:8.75,0.7515)
--(axis cs:8.75,0.697325)
--cycle;
\path [draw=white!27.843137254902!black, fill=color10, semithick]
(axis cs:9.75,0.681525)
--(axis cs:10.25,0.681525)
--(axis cs:10.25,0.73955)
--(axis cs:9.75,0.73955)
--(axis cs:9.75,0.681525)
--cycle;
\addplot [semithick, white!27.843137254902!black]
table {%
0 0.7811
0 0.7678
};
\addplot [semithick, white!27.843137254902!black]
table {%
0 0.791475
0 0.8055
};
\addplot [semithick, white!27.843137254902!black]
table {%
-0.125 0.7678
0.125 0.7678
};
\addplot [semithick, white!27.843137254902!black]
table {%
-0.125 0.8055
0.125 0.8055
};
\addplot [semithick, white!27.843137254902!black]
table {%
1 0.77975
1 0.7678
};
\addplot [semithick, white!27.843137254902!black]
table {%
1 0.78945
1 0.8004
};
\addplot [semithick, white!27.843137254902!black]
table {%
0.875 0.7678
1.125 0.7678
};
\addplot [semithick, white!27.843137254902!black]
table {%
0.875 0.8004
1.125 0.8004
};
\addplot [black, mark=diamond*, mark size=2.5, mark options={solid,fill=white!27.843137254902!black}, only marks]
table {%
1 0.7617
};
\addplot [semithick, white!27.843137254902!black]
table {%
2 0.775725
2 0.7627
};
\addplot [semithick, white!27.843137254902!black]
table {%
2 0.78645
2 0.8014
};
\addplot [semithick, white!27.843137254902!black]
table {%
1.875 0.7627
2.125 0.7627
};
\addplot [semithick, white!27.843137254902!black]
table {%
1.875 0.8014
2.125 0.8014
};
\addplot [semithick, white!27.843137254902!black]
table {%
3 0.77265
3 0.7515
};
\addplot [semithick, white!27.843137254902!black]
table {%
3 0.7872
3 0.8035
};
\addplot [semithick, white!27.843137254902!black]
table {%
2.875 0.7515
3.125 0.7515
};
\addplot [semithick, white!27.843137254902!black]
table {%
2.875 0.8035
3.125 0.8035
};
\addplot [semithick, white!27.843137254902!black]
table {%
4 0.7648
4 0.7413
};
\addplot [semithick, white!27.843137254902!black]
table {%
4 0.78235
4 0.7984
};
\addplot [semithick, white!27.843137254902!black]
table {%
3.875 0.7413
4.125 0.7413
};
\addplot [semithick, white!27.843137254902!black]
table {%
3.875 0.7984
4.125 0.7984
};
\addplot [black, mark=diamond*, mark size=2.5, mark options={solid,fill=white!27.843137254902!black}, only marks]
table {%
4 0.7383
};
\addplot [semithick, white!27.843137254902!black]
table {%
5 0.7576
5 0.7352
};
\addplot [semithick, white!27.843137254902!black]
table {%
5 0.7729
5 0.7943
};
\addplot [semithick, white!27.843137254902!black]
table {%
4.875 0.7352
5.125 0.7352
};
\addplot [semithick, white!27.843137254902!black]
table {%
4.875 0.7943
5.125 0.7943
};
\addplot [black, mark=diamond*, mark size=2.5, mark options={solid,fill=white!27.843137254902!black}, only marks]
table {%
5 0.723
5 0.8014
};
\addplot [semithick, white!27.843137254902!black]
table {%
6 0.7475
6 0.72
};
\addplot [semithick, white!27.843137254902!black]
table {%
6 0.7699
6 0.7821
};
\addplot [semithick, white!27.843137254902!black]
table {%
5.875 0.72
6.125 0.72
};
\addplot [semithick, white!27.843137254902!black]
table {%
5.875 0.7821
6.125 0.7821
};
\addplot [black, mark=diamond*, mark size=2.5, mark options={solid,fill=white!27.843137254902!black}, only marks]
table {%
6 0.6446
6 0.6395
6 0.6741
};
\addplot [semithick, white!27.843137254902!black]
table {%
7 0.74005
7 0.7077
};
\addplot [semithick, white!27.843137254902!black]
table {%
7 0.7617
7 0.7892
};
\addplot [semithick, white!27.843137254902!black]
table {%
6.875 0.7077
7.125 0.7077
};
\addplot [semithick, white!27.843137254902!black]
table {%
6.875 0.7892
7.125 0.7892
};
\addplot [black, mark=diamond*, mark size=2.5, mark options={solid,fill=white!27.843137254902!black}, only marks]
table {%
7 0.6772
7 0.6721
7 0.5937
7 0.6996
7 0.7057
7 0.6141
};
\addplot [semithick, white!27.843137254902!black]
table {%
8 0.724825
8 0.6874
};
\addplot [semithick, white!27.843137254902!black]
table {%
8 0.7637
8 0.7851
};
\addplot [semithick, white!27.843137254902!black]
table {%
7.875 0.6874
8.125 0.6874
};
\addplot [semithick, white!27.843137254902!black]
table {%
7.875 0.7851
8.125 0.7851
};
\addplot [black, mark=diamond*, mark size=2.5, mark options={solid,fill=white!27.843137254902!black}, only marks]
table {%
8 0.6548
8 0.6517
8 0.6517
8 0.6548
};
\addplot [semithick, white!27.843137254902!black]
table {%
9 0.697325
9 0.6273
};
\addplot [semithick, white!27.843137254902!black]
table {%
9 0.7515
9 0.777
};
\addplot [semithick, white!27.843137254902!black]
table {%
8.875 0.6273
9.125 0.6273
};
\addplot [semithick, white!27.843137254902!black]
table {%
8.875 0.777
9.125 0.777
};
\addplot [black, mark=diamond*, mark size=2.5, mark options={solid,fill=white!27.843137254902!black}, only marks]
table {%
9 0.5886
9 0.5713
};
\addplot [semithick, white!27.843137254902!black]
table {%
10 0.681525
10 0.5947
};
\addplot [semithick, white!27.843137254902!black]
table {%
10 0.73955
10 0.7709
};
\addplot [semithick, white!27.843137254902!black]
table {%
9.875 0.5947
10.125 0.5947
};
\addplot [semithick, white!27.843137254902!black]
table {%
9.875 0.7709
10.125 0.7709
};
\addplot [black, mark=diamond*, mark size=2.5, mark options={solid,fill=white!27.843137254902!black}, only marks]
table {%
10 0.5193
10 0.556
10 0.5509
};
\addplot [semithick, white!27.843137254902!black]
table {%
-0.25 0.78565
0.25 0.78565
};
\addplot [semithick, white!27.843137254902!black]
table {%
0.75 0.7831
1.25 0.7831
};
\addplot [semithick, white!27.843137254902!black]
table {%
1.75 0.7821
2.25 0.7821
};
\addplot [semithick, white!27.843137254902!black]
table {%
2.75 0.778
3.25 0.778
};
\addplot [semithick, white!27.843137254902!black]
table {%
3.75 0.7739
4.25 0.7739
};
\addplot [semithick, white!27.843137254902!black]
table {%
4.75 0.7658
5.25 0.7658
};
\addplot [semithick, white!27.843137254902!black]
table {%
5.75 0.7602
6.25 0.7602
};
\addplot [semithick, white!27.843137254902!black]
table {%
6.75 0.7525
7.25 0.7525
};
\addplot [semithick, white!27.843137254902!black]
table {%
7.75 0.7454
8.25 0.7454
};
\addplot [semithick, white!27.843137254902!black]
table {%
8.75 0.7312
9.25 0.7312
};
\addplot [semithick, white!27.843137254902!black]
table {%
9.75 0.71895
10.25 0.71895
};
\end{axis}

\end{tikzpicture}
    \caption{Create new BERT models by keeping $k$ layers fixed and shuffling the rest. The models are then fine-tuned again on MNLIm-sub before evaluating. Baseline: 80.14\% (dashed).}
    \label{fig:shuffle-continue-train-fixklayer-100random}
\end{figure}

Figure~\ref{fig:shuffle-continue-train-fixklayer-100random} shows what happens when we fine-tune the shuffled models for another three epochs. Remarkably, for values of $k>=5$, there are some models that reach or even surpass the baseline score. Even for $k<5$, much of the performance can be recovered. Note that 3 epochs of fine-tuning correspond to less than 2000 weight updates at a reduced learning rate, as is common when fine-tuning BERT. This stands in stark contrast to the over 1M weight updates during pre-training, and shows that most layers of BERT can be quickly \quotes{repurposed} after shuffling, which in turn indicates that they are very similar to each other.

\begin{figure}[t]
    \centering
\begin{tikzpicture}

\begin{axis}[
legend cell align={left},
legend columns = 2,
legend style={draw=white!80.0!black, anchor=south west, font=\scriptsize, anchor = north east, at={(1, 0.9)}},
height = \plotHeightSmall,
width = \linewidth,
tick align=outside,
tick pos=left,
x grid style={white!69.01960784313725!black},
xlabel={Layer},
xmin=-0.55, xmax=11.55,
xtick style={color=black},
xtick={0,1,2,3,4,5,6,7,8,9,10,11},
xticklabels={1,2,3,4,5,6,7,8,9,10,11,12},
y grid style={white!69.01960784313725!black},
ylabel={Accuracy},
ymin=0.26941, ymax=0.82,
ytick style={color=black}
]
\path [draw=black, semithick, dash pattern=on 5.550000000000001pt off 2.4000000000000004pt]
(axis cs:0,0.8014)
--(axis cs:11,0.8014);
\addplot [semithick, green!50.19607843137255!black, mark=*, mark size=2, mark options={solid}]
table {%
0 0.3289
1 0.3269
2 0.3086
3 0.2882
4 0.2882
5 0.2882
6 0.2912
7 0.3035
8 0.2974
9 0.3035
10 0.2902
11 0.3839
};
\addlegendentry{Direct eval.}
\addplot [semithick, blue, mark=square*, mark size=2, mark options={solid}]
table {%
0 0.5183
1 0.6466
2 0.6629
3 0.664
4 0.6589
5 0.498
6 0.5621
7 0.5387
8 0.5316
9 0.3574
10 0.5295
11 0.3544
};
\addlegendentry{3 more ep.}
\end{axis}

\end{tikzpicture}
    \caption{Performance of ALBERT-like models on MNLIm-sub. Baseline: 80.14\% (dashed).}
    \label{fig:albert-like-finetuned}
\end{figure}
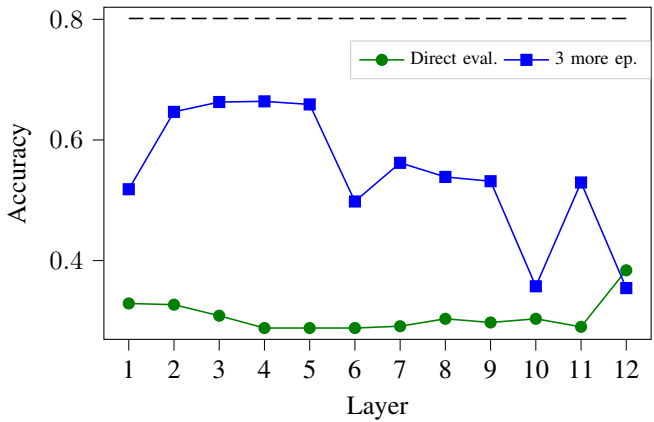

Finally, we create ALBERT-like models from the fine-tuned BERT-base by repeating each layer 12 times. Without further fine-tuning, this reduces performance to random guessing. However, if we fine-tune for another 3 epochs, performance improves significantly. The results are shown in Figure~\ref{fig:albert-like-finetuned}. Interestingly, models based on layers 2-5 seem to be best, and higher layers fail to perform well on their own.
This shows that, while BERT has an inherent tendency towards weight-sharing, its layers do still exhibit a certain hierarchical structure.

So far, the results in this section suggest that BERT has an inductive bias towards learning very similar transformations at each layer, which is why layers can be swapped and shuffled without incurring a large performance loss. We therefore hypothesize that enforcing weight sharing as in ALBERT~\cite{Lan2020ALBERT} makes this inherent tendency explicit, thus acting as an effective regularizer.
Next, we investigate where this inherent weight-sharing tendency is coming from.

\subsection{What Causes the Inductive Bias Towards Weight-Sharing?}

We now contrast the results on BERT with a simple CNN baseline trained on MNIST, where we find that swapping adjacent layers instantly reduces the classifier performance to random. 
Thus, in order to figure out which component of BERT is responsible for the strong commutatity of its layers, we train two additional CNN versions, one with layer normalization and another one with skip connections around every convolutional layer\footnote{Ablating components from BERT (e.g. skip connections) prevented convergence during training and thus, we use a simpler CNN model to analyze the behavior of the different architectural elements.}. The results for all three CNNs are shown in Figure~\ref{fig:swap-cnn-long-mnist}.
Adding layer normalization does not make the CNN more commutative. However, the CNN with skip connections shows almost no performance drop when swapping adjacent layers. These result are very similar to the ones on BERT from Figure~\ref{fig:swapping_two_bertbase}. 

\begin{figure}[t]
    \centering
\begin{tikzpicture}


\definecolor{color0}{rgb}{0.916118471592631,0.587423760636183,0.640263770354658}
\definecolor{color1}{rgb}{0.863808356178602,0.573152210740463,0.307810812811634}
\definecolor{color2}{rgb}{0.700010791112321,0.614790479109793,0.283088087205159}
\definecolor{color3}{rgb}{0.557476983073822,0.647041308839097,0.274582905586777}
\definecolor{color4}{rgb}{0.283041109678183,0.688949210907788,0.403705480982841}
\definecolor{color5}{rgb}{0.291161607362895,0.676022094321277,0.595274339910988}
\definecolor{color6}{rgb}{0.300347004618729,0.669923443048919,0.691845126957855}
\definecolor{color7}{rgb}{0.330051495547042,0.675770945828907,0.827604419705628}
\definecolor{color8}{rgb}{0.664302243915371,0.674348614874934,0.921603226420326}
\definecolor{color9}{rgb}{0.842866822558884,0.584940072236322,0.909928522149295}
\definecolor{color10}{rgb}{0.909718848041829,0.562724016816337,0.786888997621333}

\begin{axis}[
tick align=outside,
tick pos=left,
height = \plotHeightMiddle,
width = \linewidth,
x grid style={white!69.0196078431373!black},
xlabel={Number of Fixed Layers},
xmin=-0.5, xmax=9.5,
xtick style={color=black},
xtick={0,1,2,3,4,5,6,7,8,9},
xticklabels={9,8,7,6,5,4,3,2,1,0},
y grid style={white!69.0196078431373!black},
ylabel={Accuracy},
ymin=-0.01464, ymax=1.04004,
ytick style={color=black}
]
\path [draw=black, semithick, dash pattern=on 5.550000000000001pt off 2.4000000000000004pt]
(axis cs:0,0.9921)
--(axis cs:10,0.9921);

\path [draw=white!23.921568627451!black, fill=color0, semithick]
(axis cs:-0.25,0.4927)
--(axis cs:0.25,0.4927)
--(axis cs:0.25,0.7863)
--(axis cs:-0.25,0.7863)
--(axis cs:-0.25,0.4927)
--cycle;
\path [draw=white!23.921568627451!black, fill=color1, semithick]
(axis cs:0.75,0.35875)
--(axis cs:1.25,0.35875)
--(axis cs:1.25,0.7224)
--(axis cs:0.75,0.7224)
--(axis cs:0.75,0.35875)
--cycle;
\path [draw=white!23.921568627451!black, fill=color2, semithick]
(axis cs:1.75,0.292275)
--(axis cs:2.25,0.292275)
--(axis cs:2.25,0.58025)
--(axis cs:1.75,0.58025)
--(axis cs:1.75,0.292275)
--cycle;
\path [draw=white!23.921568627451!black, fill=color3, semithick]
(axis cs:2.75,0.201825)
--(axis cs:3.25,0.201825)
--(axis cs:3.25,0.410925)
--(axis cs:2.75,0.410925)
--(axis cs:2.75,0.201825)
--cycle;
\path [draw=white!23.921568627451!black, fill=color4, semithick]
(axis cs:3.75,0.149225)
--(axis cs:4.25,0.149225)
--(axis cs:4.25,0.30355)
--(axis cs:3.75,0.30355)
--(axis cs:3.75,0.149225)
--cycle;
\path [draw=white!23.921568627451!black, fill=color5, semithick]
(axis cs:4.75,0.12065)
--(axis cs:5.25,0.12065)
--(axis cs:5.25,0.216425)
--(axis cs:4.75,0.216425)
--(axis cs:4.75,0.12065)
--cycle;
\path [draw=white!23.921568627451!black, fill=color6, semithick]
(axis cs:5.75,0.11785)
--(axis cs:6.25,0.11785)
--(axis cs:6.25,0.190575)
--(axis cs:5.75,0.190575)
--(axis cs:5.75,0.11785)
--cycle;
\path [draw=white!23.921568627451!black, fill=color7, semithick]
(axis cs:6.75,0.1156)
--(axis cs:7.25,0.1156)
--(axis cs:7.25,0.1634)
--(axis cs:6.75,0.1634)
--(axis cs:6.75,0.1156)
--cycle;
\path [draw=white!23.921568627451!black, fill=color8, semithick]
(axis cs:7.75,0.104325)
--(axis cs:8.25,0.104325)
--(axis cs:8.25,0.173925)
--(axis cs:7.75,0.173925)
--(axis cs:7.75,0.104325)
--cycle;
\path [draw=white!23.921568627451!black, fill=color9, semithick]
(axis cs:8.75,0.1033)
--(axis cs:9.25,0.1033)
--(axis cs:9.25,0.148725)
--(axis cs:8.75,0.148725)
--(axis cs:8.75,0.1033)
--cycle;
\addplot [semithick, white!23.921568627451!black]
table {%
0 0.4927
0 0.2768
};
\addplot [semithick, white!23.921568627451!black]
table {%
0 0.7863
0 0.9836
};
\addplot [semithick, white!23.921568627451!black]
table {%
-0.125 0.2768
0.125 0.2768
};
\addplot [semithick, white!23.921568627451!black]
table {%
-0.125 0.9836
0.125 0.9836
};
\addplot [semithick, white!23.921568627451!black]
table {%
1 0.35875
1 0.1539
};
\addplot [semithick, white!23.921568627451!black]
table {%
1 0.7224
1 0.9921
};
\addplot [semithick, white!23.921568627451!black]
table {%
0.875 0.1539
1.125 0.1539
};
\addplot [semithick, white!23.921568627451!black]
table {%
0.875 0.9921
1.125 0.9921
};
\addplot [semithick, white!23.921568627451!black]
table {%
2 0.292275
2 0.0731
};
\addplot [semithick, white!23.921568627451!black]
table {%
2 0.58025
2 0.9921
};
\addplot [semithick, white!23.921568627451!black]
table {%
1.875 0.0731
2.125 0.0731
};
\addplot [semithick, white!23.921568627451!black]
table {%
1.875 0.9921
2.125 0.9921
};
\addplot [semithick, white!23.921568627451!black]
table {%
3 0.201825
3 0.0969
};
\addplot [semithick, white!23.921568627451!black]
table {%
3 0.410925
3 0.6826
};
\addplot [semithick, white!23.921568627451!black]
table {%
2.875 0.0969
3.125 0.0969
};
\addplot [semithick, white!23.921568627451!black]
table {%
2.875 0.6826
3.125 0.6826
};
\addplot [black, mark=diamond*, mark size=2.5, mark options={solid,fill=white!23.921568627451!black}, only marks]
table {%
3 0.9066
3 0.9921
3 0.948
3 0.8237
};
\addplot [semithick, white!23.921568627451!black]
table {%
4 0.149225
4 0.0903
};
\addplot [semithick, white!23.921568627451!black]
table {%
4 0.30355
4 0.5301
};
\addplot [semithick, white!23.921568627451!black]
table {%
3.875 0.0903
4.125 0.0903
};
\addplot [semithick, white!23.921568627451!black]
table {%
3.875 0.5301
4.125 0.5301
};
\addplot [black, mark=diamond*, mark size=2.5, mark options={solid,fill=white!23.921568627451!black}, only marks]
table {%
4 0.9738
4 0.6032
4 0.6085
4 0.9193
};
\addplot [semithick, white!23.921568627451!black]
table {%
5 0.12065
5 0.0458
};
\addplot [semithick, white!23.921568627451!black]
table {%
5 0.216425
5 0.359
};
\addplot [semithick, white!23.921568627451!black]
table {%
4.875 0.0458
5.125 0.0458
};
\addplot [semithick, white!23.921568627451!black]
table {%
4.875 0.359
5.125 0.359
};
\addplot [black, mark=diamond*, mark size=2.5, mark options={solid,fill=white!23.921568627451!black}, only marks]
table {%
5 0.38
5 0.5016
5 0.8105
5 0.4845
5 0.3792
5 0.3944
5 0.6407
};
\addplot [semithick, white!23.921568627451!black]
table {%
6 0.11785
6 0.067
};
\addplot [semithick, white!23.921568627451!black]
table {%
6 0.190575
6 0.2791
};
\addplot [semithick, white!23.921568627451!black]
table {%
5.875 0.067
6.125 0.067
};
\addplot [semithick, white!23.921568627451!black]
table {%
5.875 0.2791
6.125 0.2791
};
\addplot [black, mark=diamond*, mark size=2.5, mark options={solid,fill=white!23.921568627451!black}, only marks]
table {%
6 0.3219
6 0.4291
6 0.336
};
\addplot [semithick, white!23.921568627451!black]
table {%
7 0.1156
7 0.0668
};
\addplot [semithick, white!23.921568627451!black]
table {%
7 0.1634
7 0.2239
};
\addplot [semithick, white!23.921568627451!black]
table {%
6.875 0.0668
7.125 0.0668
};
\addplot [semithick, white!23.921568627451!black]
table {%
6.875 0.2239
7.125 0.2239
};
\addplot [black, mark=diamond*, mark size=2.5, mark options={solid,fill=white!23.921568627451!black}, only marks]
table {%
7 0.2437
7 0.3656
7 0.2582
7 0.3299
7 0.2526
7 0.2939
7 0.3284
};
\addplot [semithick, white!23.921568627451!black]
table {%
8 0.104325
8 0.0333
};
\addplot [semithick, white!23.921568627451!black]
table {%
8 0.173925
8 0.2564
};
\addplot [semithick, white!23.921568627451!black]
table {%
7.875 0.0333
8.125 0.0333
};
\addplot [semithick, white!23.921568627451!black]
table {%
7.875 0.2564
8.125 0.2564
};
\addplot [black, mark=diamond*, mark size=2.5, mark options={solid,fill=white!23.921568627451!black}, only marks]
table {%
8 0.2874
8 0.2987
8 0.2808
8 0.3038
};
\addplot [semithick, white!23.921568627451!black]
table {%
9 0.1033
9 0.0535
};
\addplot [semithick, white!23.921568627451!black]
table {%
9 0.148725
9 0.2144
};
\addplot [semithick, white!23.921568627451!black]
table {%
8.875 0.0535
9.125 0.0535
};
\addplot [semithick, white!23.921568627451!black]
table {%
8.875 0.2144
9.125 0.2144
};
\addplot [black, mark=diamond*, mark size=2.5, mark options={solid,fill=white!23.921568627451!black}, only marks]
table {%
9 0.2197
9 0.2409
};
\addplot [semithick, white!23.921568627451!black]
table {%
-0.25 0.6165
0.25 0.6165
};
\addplot [semithick, white!23.921568627451!black]
table {%
0.75 0.51045
1.25 0.51045
};
\addplot [semithick, white!23.921568627451!black]
table {%
1.75 0.37065
2.25 0.37065
};
\addplot [semithick, white!23.921568627451!black]
table {%
2.75 0.28145
3.25 0.28145
};
\addplot [semithick, white!23.921568627451!black]
table {%
3.75 0.1936
4.25 0.1936
};
\addplot [semithick, white!23.921568627451!black]
table {%
4.75 0.1597
5.25 0.1597
};
\addplot [semithick, white!23.921568627451!black]
table {%
5.75 0.1448
6.25 0.1448
};
\addplot [semithick, white!23.921568627451!black]
table {%
6.75 0.1473
7.25 0.1473
};
\addplot [semithick, white!23.921568627451!black]
table {%
7.75 0.133
8.25 0.133
};
\addplot [semithick, white!23.921568627451!black]
table {%
8.75 0.11945
9.25 0.11945
};
\end{axis}

\end{tikzpicture}
    \caption{Creating new skip-connection CNN models by keeping $k$ layers fixed and shuffling the rest. The resulting models are evaluated without further fine-tuning. Baseline: 99.1\% (dashed).}
    \label{fig:shuffle-direct-eval-cnn-long-ln-sc-fixklayer-100randomt}
\end{figure}
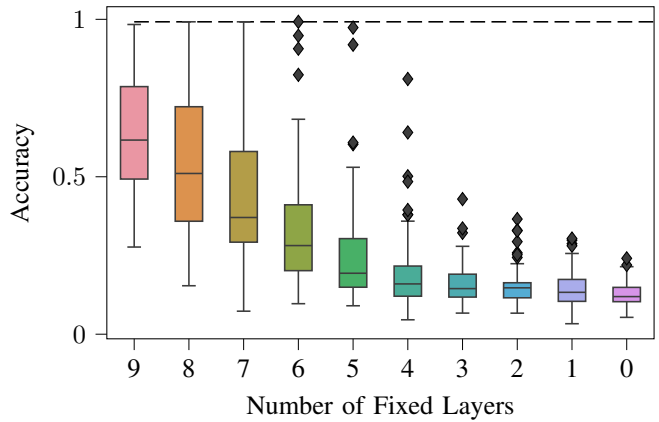

Analogously to the previous section, we also investigate how the CNN behaves when more than two layers are shuffled. The results are shown in Figure~\ref{fig:shuffle-direct-eval-cnn-long-ln-sc-fixklayer-100randomt}. 
Direct comparisons to the results on BERT are difficult due to the differences in training task and architectures. However, the figures clearly show that layers of models with skip connections are highly commutative, especially when compared to baselines without skip connections. 
These results suggests that much of the inherent weight-sharing bias of BERT comes from the skip connections in the FFN and Attention blocks.

\begin{figure}[t]
    \centering
\begin{tikzpicture}

\begin{axis}[
legend cell align={left},
legend style={at={(0.91,0.55)}, anchor=east, draw=white!80.0!black, font=\scriptsize},
legend columns = 3,
tick align=outside,
tick pos=left,
height = \mainPlotHeight,
width = \linewidth,
x grid style={white!69.01960784313725!black},
xlabel={Swapped Layers},
xmin=-0.45, xmax=9.45,
xtick style={color=black},
xtick={0,1,2,3,4,5,6,7,8,9},
xticklabels={{(2,3)},{(3,4)},{(4,5)},{(5,6)},{(6,7)},{(7,8)},{(8,9)},{(9,10)},{(10,11)},{(11,12)}},
xticklabel style={font=\small, rotate=45},
y grid style={white!69.01960784313725!black},
ytick={0, 0.25, 0.5, 0.75, 1},
ylabel={Accuracy},
ymin=-0.03432, ymax=1.1,
ytick style={color=black}
]

\path [draw=green!50.19607843137255!black, semithick, dash pattern=on 5.550000000000001pt off 2.4000000000000004pt]
(axis cs:0,0.9910)
--(axis cs:9,0.9910);
\path [draw=blue, semithick, dash pattern=on 5.550000000000001pt off 2.4000000000000004pt]
(axis cs:0,0.9921)
--(axis cs:9,0.9921);
\path [draw=black, semithick, dash pattern=on 5.550000000000001pt off 2.4000000000000004pt]
(axis cs:0,0.9921)
--(axis cs:9,0.9921);

\addplot [semithick, green!50.19607843137255!black, mark=*, mark size=2, mark options={solid}]
table {%
0 0.2133
1 0.1104
2 0.178
3 0.0567
4 0.0691
5 0.0938
6 0.0883
7 0.0529
8 0.2477
9 0.1521
};
\addlegendentry{CNN}
\addplot [semithick, blue, mark=square*, mark size=2, mark options={solid}]
table {%
0 0.1261
1 0.0925
2 0.0803
3 0.0543
4 0.2802
5 0.0305
6 0.0735
7 0.0143
8 0.1259
9 0.1021
};
\addlegendentry{CNN-LN}
\addplot [semithick, black, mark=triangle*, mark size=2, mark options={solid}]
table {%
0 0.928
1 0.958
2 0.9233
3 0.9744
4 0.9738
5 0.9748
6 0.9836
7 0.9271
8 0.9867
9 0.9805
};
\addlegendentry{CNN-LN-SC}
\end{axis}

\end{tikzpicture}
    \caption{Swapping adjacent layers of CNNs trained on MNIST. LN stands for layer normalization and SC for skip connections. Baseline: 99.1\% (dashed).}
    \label{fig:swap-cnn-long-mnist}
\end{figure}
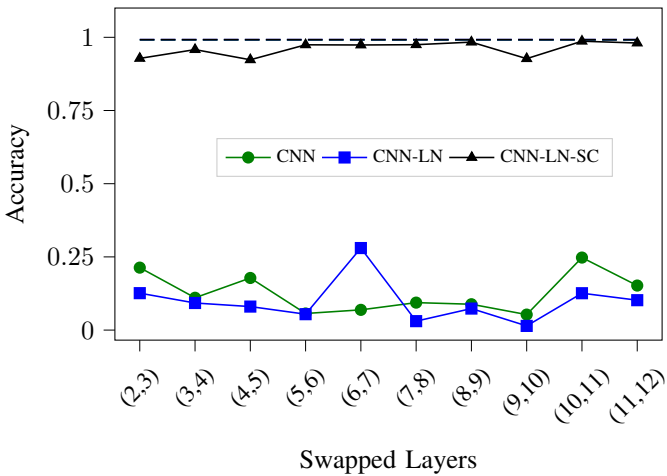

Our findings in this section suggest that feature extraction in BERT is not strictly hierarchical, but happens incrementally, and that it is not the \quotes{early layers} per-se that extract low-level features, but low-level features are simply extracted first, regardless of the order in which the layers are applied. 
We believe that this is partly due to the commutativity of BERT's layers, which seems to be caused mostly by the use of skip connections.

\section{Conclusion}

In this work we propose a tool to measure non-linearity in transformers which takes into account the geometry of the embedding space. With this, we find that the non-linearity introduced by the MLPs and SA-FF blocks follow the same U-shaped pattern.
Further, we find that although FFN blocks contain approximately two thirds of the parameters, attention has a considerably stronger impact on the performance of the model. 
Finally, we investigate how BERT's layers interact with each other and observe that features are extracted in a fuzzy or incremental manner, with the layers being very similar to each other. By comparing BERT to simple CNNs we find that the commutativity of the layers is induced by the skip connections.
This inductive bias towards weight sharing and recurrence provides an explanation for the strong performance of weight-shared models like ALBERT or the Universal Transformer.

We hope that this work will motivate further investigation of architectural elements of transformers other than self-attention. We believe a more efficient use of the parameters contained in the FFNs is possible, and we expect that further research will find a better way of exploiting the representational power of those parameters. 
Although we find that skip connections make adjacent layers commutative, the role of self-attention in enabling recurrence over layers remains unclear. Since self-attention is an input-dependent operation, it provides additional flexibility to the model that may further enforce the inductive bias towards weight-sharing. Hence, we deem the investigation of weight-shared architectures an exciting direction for future work.

\bibliographystyle{plain}
\bibliography{bibliography}

\end{document}